# Characterizing machine learning process: A maturity framework

Rama Akkiraju, Vibha Sinha, Anbang Xu, Jalal Mahmud, Pritam Gundecha, Zhe Liu, Xiaotong Liu, John Schumacher

IBM Watson, IBM Almaden Research Center, San Jose, California, USA

{akkiraju, vibha.sinha, anbangxu, jumahmud, psgundec, liuzh, Xiaotong.Liu, jfs}@us.ibm.com

*Abstract*—Academic literature on machine learning modeling fails to address how to make machine learning models work for enterprises. For example, existing machine learning processes cannot address how to define business use cases for an AI application, how to convert business requirements from offering managers into data requirements for data scientists, and how to continuously improve AI applications in term of accuracy and fairness, how to customize general purpose machine learning models with industry, domain, and use case specific data to make them more accurate for specific situations etc. Making AI work for enterprises requires special considerations, tools, methods and processes. In this paper we present a maturity framework for machine learning model lifecycle management for enterprises. Our framework is a re-interpretation of the software Capability Maturity Model (CMM) for machine learning model development process. We present a set of best practices from authors' personal experience of building large scale real-world machine learning models to help organizations achieve higher levels of maturity independent of their starting point.

*Keywords—machine learning models, maturity model, maturity framework, AI model life cycle management.*

## I. Introduction

Software and Services development has gone through various phases of maturity in the past few decades. The community has evolved lifecycle management theories and practices to disseminate best practices to developers, companies and consultants alike. For example, in software field, Software Development Life Cycle (SDLC) Management, capability maturity models (CMM) Application Life Cycle Management (ALM), Product Life Cycle Management (PLM) models prescribe systematic theories and practical guidance for developing products in general, and software products in particular. Information Technology Infrastructure Library (ITIL) organization presents a set of detailed practices for IT Services management (ITSM) by aligning IT services with business objectives. All these practices provide useful guidance for developers in systematically building software and services assets. However, these methods fall short in managing a new breed of software services being developed rapidly in the industry. These are software services built with machine learnt models.

We are well into the era of Artificial Intelligence (AI), spurred by algorithmic, and computational advances, the availability of the latest algorithms in various software libraries, Cloud technologies, and the desire of companies to unleash insights from the vast amounts of untapped unstructured data lying in their enterprises. Companies are actively exploring and deploying trial versions of AI-enabled applications such as chat bots, personal digital assistants, doctors' assistants, radiology assistants, legal assistants, health and wellness coaches in their enterprises. Powering these applications are the AI building block services such as conversation enabling service, speech-to-text and text to speech, image recognition service, language translation and natural language understanding services that detect entities, relations, keywords, concepts, sentiments and emotions in text. Several of these services are machine learnt, if not all. As more and more machine learnt services make their way into software applications, which themselves are part of business processes, robust life cycle management of these machine learnt models becomes critical for ensuring the integrity of business processes that rely on them. We argue that two reasons necessitate a new maturity framework for machine learning models. First, the lifecycle of machine learning models is significantly different from that of the traditional software and therefore a reinterpretation of the software capability maturity model (CMM) maturity framework for building and managing the lifecycle of machine learning models is called for. Second, building machine learning models that work for enterprises requires solutions to a very different set of problems than the academic literature on machine learning typically focuses on. We explain these two reasons below a bit more in detail.

### A. Traditional Software Development Vs. Machine Learning Model Development

While traditional software applications are deterministic, machine learning models are probabilistic. Machine learning models learn from data. They need to be trained while traditional software applications are programmed to behave as per the requirements and specifications. As a result, traditional software applications are always accurate barring defects, whereas machine learning models typically need multiple iterations of improvements to achieve acceptable levels of accuracy, and it may or may not be possible to achieve 100% accuracy. Data in traditional software applications tends to be transactional in nature and mostly of structured type whereas data for machine learning models can be structured, or unstructured. Unstructured data can further come in multiple forms such as text, audio, video and images. In addition, data management in machine learning pipeline has multiple stages, namely data acquisition, data annotation, data preparation, data quality checking, data sampling, data augmentation steps – each involving their own life cycles thereby necessitating a whole new set of processes and tools. Machine learning models have to deal with fairness, trust, transparency, explainability that traditional software doesn't have to deal with. Machine learning pipeline has a whole new set of roles such as data managers, data annotators, data scientists, fairness testers etc. in addition to traditional software engineering roles. While one has to deal with code versioning and code diff functions in traditional software application development, machine learning models bring interesting twists with training data and testing data diffs and model diffs. A full version of compare and contrast is the sole subject of a different paper under preparation.

All these new aspects in machine learning model lifecycle need explication, disciplined management and optimization lest organizations end up with chaotic, poor quality models thereby leaving a trail of dissatisfied customers.

*B. Making Machine Learning and AI work for Enterprises*

Traditional machine learning process in academia does not consider new challenges faced by enterprises when deploying machine learning models. For example, existing machine learning process cannot address how to define business use cases for an AI application, how to convert business requirements from offering managers into data requirements for data scientists, and how to continuously improve AI applications in term of accuracy and fairness, how to customize general purpose machine learning models with industry, domain, and use case specific data to make them more accurate for specific situations etc. Making AI work for enterprises requires special considerations, tools, methods and processes. This further necessitates a new maturity framework for machine learning models.

To address these problems, based on our own experience of building practical, large-scale, real-world, machine learning models, we present a new interpretation of CMM maturity framework for managing the lifecycle of machine learnt models. To the best of our knowledge, this is the first of its kind.

In this paper we use machine learning model and AI model synonymously, although we understand that machine learning models are only a type of AI models.

## II. RELATED WORK

Our work is related to software maturity model [1], Big data maturity models [3, 4, 6, 7, 8], and knowledge discovery process [10].

Humphrey proposed capability maturity model (CMM) for Software [1]. He described five levels of process maturity for Software: *initial*, *repeatable*, *defined*, *managed*, *optimizing*. An organization's maturity is considered *initial* when there is no control of the process and no orderly progress of process improvement is possible. An organization can reach *repeatable* level when it has achieved a stable process with repeatable level of statistical control by initiating rigid project management of commitments, cost, schedule and changes. *Defined* level can be attained when the organization has defined the process to ensure consistent implementation and provide a basis for better understanding of the process. An organization attains a *managed* level when it has initiated comprehensive process measurements beyond those of cost and schedule performance. An organization reaches *optimizing* level when the organization has a foundation for continuous improvement and optimization of the process. Our work is inspired by such process maturity definitions. In our work, we propose a set of required processes for organizations building machine learning models.

Big data maturity has been proposed for organizations to track their progress and to identify relevant initiatives [2].

According to [3], big data maturity is not simply about having technology in place to deal with high volumes of data. It also involves the processes of "building an ecosystem that includes technologies, data management, analytics, governance". A number of big data maturity management (BDMM) models have been proposed. Among the several existing ones, IBM's Big Data and Analytics Maturity Model [4], TDWI Big Data Maturity Model [3], EMC's Big Data Business Model Maturity Index [5], Hortonworks Big Data Maturity Model [6], Booz & Company's BDMM [7], and Radcliffe's BDMM [8] are the popular ones. Most of these BDMMs adopt maturity grids with 4 to 6 different phases (e.g. ad-hoc, foundational, competitive, differentiating, breakaway) to assess the big data maturity management in various aspects of organizational areas such as business strategy, information, analytics, culture and operational execution, architecture, governance, etc. More detailed evaluation of different BDMM can be found in [2] and [9].

Han et al. [10] described a typical knowledge discovery process consisting of the following stages: (1) **data cleaning** to remove noise and inconsistent data; (2) **data integration**, where multiple data sources may be combined; (3) **data selection**, where data relevant to the analysis task are retrieved from the database; (4) **data transformation**, where data are transformed and consolidated into forms appropriate for mining; (5) **data mining**, where intelligent methods are applied to extract data patterns; (6) **pattern evaluation** to identify the truly interesting patterns; (7) **knowledge presentation**, which presents mined knowledge to users. Our work aims to provide organizations and practitioners with a comprehensive maturity level assessment of the major components in the machine learning pipeline. To the best of our knowledge, our work is the first maturity framework to characterize a machine learning process.

## III. MACHINE LEARNING MODEL LIFECYCLE

In this section we describe the AI Service development lifecycle, along with roles involved in each. AI lifecycle include: data pipeline, feature pipeline, train pipeline, test pipeline, deployment pipeline, and continuous improvement pipeline. Each step is an iterative and requires continuous improvements in itself. This iterative process is illustrated in Figure 1. A brief introduction to each step is given in this section. The sections that follow provide deep-dives and maturity assessment questionnaire.

1) <u>Model Goal Setting and Offering Management:</u> An offering manager kicks off the AI model development process by setting goals for the AI model i.e., what must it be good at, creates test cases and minimum required thresholds upon which the models' quality and runtime performance targets are to be measured. This person also defines thresholds for model competitiveness and the associated levels. An offering manager must set goals for an AI model considering the current state as well as achievable levels with stretch targets. The goals must apply not only the model quality and runtime metrics but also to the process by which the models are built so that the outcomes are predictable, consistent and repeatable. After the initial model is built, assessed and certified for deployment, an offering manager must monitor the quality and runtime performance of the model across various versions of the model to ensure model alignment with goals, and targets. This person is also responsible for ensuring that the model is free of undesirable biases, fair, transparent and keeps track of the client service-level agreement (SLAs) needs and model performance across various versions of the model. This person or team must also estimate the data budgets and make a business case to the Finance department to procure the needed funds to acquire data, if external data is needed or to other divisions within the company if data is available internally to build a model. Often the importance and time it takes to get approvals to acquire data either internally or from external sources is underestimated, leading to frustrations for the data science team being readied for training and testing.

2) *Content Management Strategy*: A content manager is responsible for proactively identifying suitable training data sources from public and private legal sources, checking the legality of data, establishing governance process around data, data vendor contract negotiations, pricing, data budget management and data lineage management.

3) *Data Pipeline*: Data collection and preparation is a key step in training an AI model. In this step, an AI Service Data Lead leads the efforts around data collection and labeled data preparation. The model needs to see enough instances of each kind that you are trying to detect/predict. For example, a Sentiment Analyzer service needs to see enough instances of positive, negative and neutral sentiment samples in order to learn to classify them correctly. This stage of data collection and ground truth preparation involves many activities such as identifying right type of data in right distributions, sampling the data so as to guide the model performance, enriching the data via labeling, storing the lineage of the data, checking the quality of the labeled and prepared data, establishing specific metrics for measuring the quality of the data, storing and analyzing the data. This step may also involve augmenting the training data via data synthesis techniques or with adversarial examples to enhance the robustness of models. Each step is iterative in itself and goes through multiple iterations before the data is readied for training.

4) *Feature Preparation*: This step involves preparing the features from the collected data to initiate the training models. The actual preparation steps depend on the type of AI service being developed. Figure 1 shows the preparatory steps involved in text processing and audio signal processing for building natural language understanding (NLU) and speech-to-text type of services. Typically, these include, developing tokenizers, sentence segmentation capabilities, part-of-speech taggers, lemmatization, syntactic parsing capabilities etc. In the case of audio data, these things include developing phonetic dictionaries, text normalizers etc. These assets and services once prepared are then used in training algorithms. Typically, a Training Lead works closely with the Data Lead to prepare these assets.

5) *Model Training*: A Training Lead leads this activity. A Training Lead makes decisions about what algorithms to experiment with the prepared data and the feature assets that are prepared. This includes making decisions about what frameworks to use (TensorFlow/Pytorch/Keras etc.), if neural nets are involved, how many hidden layers and the specific activation functions at each layer etc. A Training Lead then trains the models, after making the train/dev/test set splits on labeled data and runs multiple experiments before finally making the model selection. Throughout the training process, Training Lead makes many decisions on the various hyper parameters and strives to optimize the network/architecture of the training algorithm to achieve best results. A Training Lead also conducts error analysis on failed training and dev/cross-validation cases and optimizes the model to reduce those errors. A Training Lead does not have access to the test cases.

6) *Testing and Benchmarking*: A Test Lead leads the testing and benchmarking activity. Finalized model is tested against multiple datasets that are collected. The model is also tested against various competitor services, if accessible, and applicable. Comparing the quality and run-time performance of the model with competitor's services and all known competing AI models to establish its quality for each model version is a critical aspect of testing phase. As noted earlier, a test lead is also responsible for conducting detailed and thorough error analysis on the failed test cases and sharing the observations and patterns with the Training Lead so as to help improve the AI model in future iterations.

7) *Model Deployment:* This is the step where critical decisions are made by the Deployment Lead on the deployment configuration of the model. In Software-as-a-Service (SaaS) services, this often involves, infrastructure components, memory, disk, CPUs/GPUs, and number of pods needed based on the expected demand. Very often as part of deployment, significant engineering might be required to make the feature extraction steps production-grade and wrap the trained model into a software package that can be invoked from the larger business application.

8) *AI Operations Management*: Any AI Service's lifecycle hardly ends when the first model is deployed for the first time by following the steps described above. Each AI model has to continuously improve overtime by learning from the mistakes it makes. With each iteration, with each feedback loop, with each new model version, the model continuously evolves. Managing these iterations that lead to continuous learning of AI services is what we call as AI Operations, and is a joint activity between the operations, data and training team.

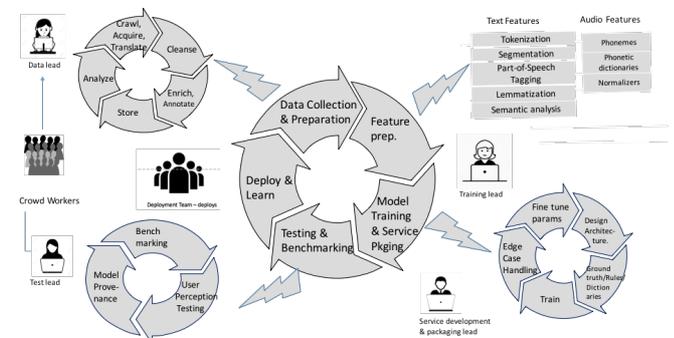

Fig. 1. AI model lifecycle

Deployment team is responsible for logging the payloads of AI models, and managing the governance of payload data with help of Data Lead. During continuous improvement cycle, the new incoming data is included by Train Lead to re-do training process and prepare a model that is more accurate for the data it is being used for. The payload data is also used to detect and address aspects such as biases, errors, model drifts, misalignments and explainability.

In the following sections we elaborate on each of these pipeline stages. In Appendix A we present a small snippet of our maturity framework. A more detailed maturity framework could not be attached due to space limitations but will be made available via company website.

## IV. Data Pipeline

Given input data X, A machine learning model approximates a mapping function f to predict an output value y such that $f(X) = y$. Training machine learning models is a data intensive effort. Training data must have enough representation of the world that the model wants to approximate. Real-world data is often messy and must be cleaned and prepared to make it usable for training AI models. Recent studies show that curating data contributes to more than half the time in the lifecycle management of AI models [11]. Since data plays a pivotal role in AI, managing the data pipeline effectively, and aligning data curation efforts with the business goals and requirements can be key differentiators for organizations. Below, we describe some strategies for managing one's data pipelines effectively.

- Define Data Requirements According to Business Needs: Mature organizations aspiring to produce high quality AI models start with defining goals for their AI models. A model Offering Lead must first define the scope, purpose and expected minimum quality thresholds for an AI model. In organizations just starting with machine learning, this strategic job is left to data scientists responsible for training. While data scientists do their best to build a good model, it is not their job or role to define what it must be good at. For example, asking data scientists to 'build a world-class face recognition AI model' is too broad and vague. A more specific and focused goal would look like this: 'build a face recognition service that can detect male, female genders, *these* specified age groups, and *these* specified subset of races, and ethnicities, which are defined in the requirements document (the requirements document may point to a more specific taxonomy of races and ethnicities to be detected from a neutral entity such as the United Nations Race and Ethnicity taxonomy)) with at least 90% accuracy on '*these*' given specific test datasets' where 'these' test datasets were carefully crafted by the offering management team to have an even distribution of all the genders, age groups, specific races, and ethnicities for which the model is supposed do well. That is a specific, focused and measurable goal that a data scientist can build a model for. Such a focused goal is also non-disputable. If the business purpose and goal is not clear, organizations have to deal with poor performance and unfairness claims once the model goes into production where users may complain that the face recognition is biased and doesn't recognize faces of certain races and ethnicities. Such a specific goal also sets specific objectives for data and training leads in collecting the right kind of data and setting right type of train, and dev splits respectively while building the model. This way, instead of shooting in the dark, an organization managing a mature data pipeline can convert high-level business goals (e.g. target industries, domains, scenario and etc.) into specific data requirements.

- Define a Data Acquisition Strategy: A mature data pipeline should be able to consider the time and cost of data curation and correlate and quantify the performance gains of AI models with the curated data. This way, an organization can justify the data curation efforts while maximizing performance gains for their AI solutions.

- Apply Data Selection to Select Suitable Training Data: The goal of data selection is to select representative, unbiased and diverse data. This is a funneling process. Data cleansing and data selection both reduce data as the result of processing. Therefore, in order to achieve desired quantities of representative data, organizations may have to be prepared to collect more data than they may end up using. If data selection is not done, on the other hand, i) models may end up with undesirable biases as proper representation may not be achieved ii) organizations may have to pay for labeling data that may or may not be useful, adding to the costs and iii) too much unselected data may unnecessarily add to the processing time and computational capacity requirements of the machine learning process. Therefore, it is critical to apply appropriate sampling techniques in order to generate quality training data sets in reasonable sizes.

- Create Data Annotation Guidelines to Achieve Consistency with Data Labeling: In general, the more the available annotated data, the better the model performs. However, labeling data can be difficult and expensive. To deliver high-quality annotated data in an efficient way, an organization should consider the following three aspects: (i) create unambiguous definitions for terms, prepare clear annotation guidelines and continuously refine the guidelines and definitions with user feedback. A mature pipeline should support a rapid feedback loop between data scientists and data annotators, (ii) use a combination of internal team of annotators and external crowd workers to get data annotated at scale, (iii) use machine learning to pre-annotate data that human annotators can validate. This can greatly speed up the human annotation process.

- Augment Data using Synthetic Techniques as Applicable: In machine learning algorithms, there is often a need to synthetically augment data, to increase the amount of training data to cover scenarios where real data might be difficult to get by. For example, in the case of audio data for training a Speech-to-Text model, a given set of audio files can be augmented by superposition of noise tracks, echoes, reverberations etc. Also, rate, pitch modulation can be performed on audio files to synthesize additional data. In the case of image recognition modeling, an image can be tilted, rotated, and colors changed to generate additional training data. As a best practice, we recommend organizations to have a strategy and develop a pipeline for data augmentation and align the augmented data requirements with specific business needs. This requires conducting different training experiments using a combination of real and synthetic data, and measuring the impact of each experiment before deciding what model to deploy.

- Standardize and Automate Quality Check Procedure: Data quality must be checked throughout the data pipeline. Data cannot be used to train/test an AI module until its quality is assessed and certified, as the quality of data from different sources and vendors can vary. To assess data quality, organizations must define metrics for measuring the quality of different types of data such as text, audio files, video files and images. Standardizing data quality metrics becomes an essential step to ensuring the quality of the data pipeline. For example, the quality of speech audio files can be measured using metrics such as Signal-to-noise-ratio (SNR). Other metrics such as cross-talk, silence detection, can be defined to measure audio file quality.

- Maintain Data Lineage: Data lineage deals with tracking the lifecycle of data artifacts used in machine learning models. Most organizations building machine learning models that we have seen don't yet have a disciplined process for maintaining data lineage and preserving all transformations data might go through. (i) Data is collected from multiple sources and often data licensing terms are ignored if noticed at all and lineage is not kept. This can lead to compliance and regulatory violations. Therefore, it is important to ensure the legality of data and to acquire proper licenses for all data that is acquired. (ii) Data is often obtained by whatever means, stored in various places including data scientists' personal laptops. Worse still, we have seen instances of training data being lost when laptops were upgraded or servers decommissioned leaving machine learning models in production in the lurch with no good path for continuous improvement. Such problems can be fixed by creating a data/content manager role and point person in organization. That data manager should be given the overall responsibility to ensure the integrity of data that is acquired from whatever source. All data acquired must be stored and versioned. (iii) Raw data is collected, it is cleansed and sampled, part of it goes through annotation process, synthetic data is added, various quality checks are applied. A lot of this work is done through ad-hoc scripts, and much of the pipeline is not maintained as a replicable process. Tools and techniques have to be brought to bear to capture the various transformations data goes through. (iv) Among other concerns associated with managing the data lineage for machine learning models is keeping the association between training data and a trained model. Each model should be able to trace back in its lifecycle what datasets it is trained with and similarly, content owners should have full visibility into all the consumers of each dataset, be it human consumers or machine learning models.

- Govern Data: Last but not the least, data acquired from different sources is subject to different access and cleansing restrictions. Training data must be carefully governed for access and the training data store must be kept compliant at all times.

## V. FEATURE PIPELINE

The success or failure of the machine learning algorithms is intimately tied to how the data is represented [16]. In this section, we present some best practices for managing the feature pipeline:

- Keep Your Options Open during Feature Selection: Researchers have explored different types of training algorithms that aim to exploit different types of feature representations. These feature representations can be grouped into 3 types, (1) raw-features (2) expert-designed features and (3) latent-features. Characters, pixels or audio waves are prime example of raw features [13, 14, 15]. Raw features require minimal pre-processing and transformations to data before being fed to the training algorithms. From engineering point-of-view, it has resulted in much simpler training and testing pipelines. However, this comes at the cost of need for large amounts of data for training. Other extreme to using raw features is using expert-designed features. Experts often bring domain knowledge to create these features. However, applying learning from one domain to other is often the Achilles heel for such algorithms. The over-dependence on expert users is often seen as a limitation in terms of time and cost. In the last decade, in particular with image and speech applications, state-of-the-art models have often used lower-level features than expert-level features. Recent advancement of deep-learning algorithms made a consistent case for third type of features known as latent-features. These features typically come from unsupervised pre-trained models. Intuitively, these features compress the high-quality information that goes beyond explicitly created features. Success of Word embeddings is a primary example of usefulness of latent features [17]. Recent advancement in GPU technologies fueled the possibility of training complex unsupervised models at a much faster-rate. Hence, unsupervised deep-learning based techniques are consistently providing much better latent features in varieties of applications that deal with texts, audios, images, and videos [12]. The main drawback of these features is that it's very hard to explain them. Hence, building the explainable model using latent features is an open research problem. A mature organization implementing machine learning pipeline should always keep the option of using all types of features and be aware of which features make more sense for a given task.

- Understand Performance Tradeoffs with Feature Processing: If feature-pipeline has to support real-world applications, then often response time of the model in production environment becomes a bottleneck in addition to the effectiveness of features. Hence, understanding the trade-offs between response times and model quality is necessary. Since most of these trade-offs are influenced by the available training datasets at the time, these trade-offs need be revisited when underlying datasets, training algorithms or requirements change significantly. To better generalize a machine learning service, organizations often collect datasets from various sources. Features are key to understanding the differences between these sources. Since collecting high-quality datasets is costly, powerful feature analysis provide clues on when to collect and how to diversify data for the training algorithms.

- Master the Art of Feature Representation: Preparing features for a given task often requires creativity. Many-a-times organization needs task-specific features to build the best model. For example, in text analysis, it's important to pay attention to how sentences are getting tokenized. Successful tokenizer segments emoticons, slangs, abbreviations to improve the overall perception of the sentiment analysis system. Organizations often need to be flexible to modify or even re-write the tokenizer to keep the task specific features. Similarly, for effective speech recognition system, creating language or even dialect specific phonetic dictionaries have shown to have better generalization with less amount of labeled data.

## VI. TRAIN PIPELINE

We present some best practices from our own experience of training large scale AI models deployed to production.

- From Experimentation to Production: Design your Compute Stategy: The train step in an AI project often starts with a single data scientist working on developing a model that learns the input and output relationship from the training data. In quest of implementing the best model, the data scientist experiments with multiple algorithms, frameworks, and configurations. While, initially, it might be sufficient to run these experiments on a local machine or couple of servers, very soon the number of experiments that need to be executed starts getting constrainted by available compute. Furthermore, often special compute is required for running specific machine learning algorithms e.g. for deep learning GPUs are preferred. Speech training requires large amount of storage when compared to storage required for running training on text data. Hence, a scalable infrastructure strategy is needed to support training needs. It is better to plan for such compute needs as soon as the initial experiments and approach shows promise.

- Data and Model Versioning for Efficient Collaboration and Experiementation: As the initial train experiments start showing promise, the data science team also grows. In order to support collaboration, coordination and reuse in a growing team version management of models become imperative. However, it is no longer just train and feature extraction code that needs to be versioned, but also the training data, along with experiment settings so any of the train experiments can be reproduced.

- Modularizing Train Code: Modularity of train code, so it becomes easy to plug in different components, is another productivity booster. A data scientist might have started off with a monolithic piece of code where data pre-processing, feature engineering, training code are all inter-twined. However, this soon becomes a problem as data science team would need to experiment with different machine learning approaches, different features, different data pre-processing steps, with different team members focusing on different pieces, and different frameworks being used for each.

- *Plan for Train to Serve Handoff Management*: While data scientists focus on building the most accurate model, the engineering team focuses on the non functional aspects such as run-time performance, capacity planning, and scaling approach. Often at this step the serve and train pipelines start differing. This is because data scientists prefer flexibility in how they process data and extract features because of which they might rely on custom ad-hoc scripts, while engineering team would need robust code optimized for runtime performance. What further accentuates this problem is also the fact that train is an offline process, and test is online. Long times to productize an AI model is a big challenge many AI projects face. As organizations mature there is increased demand for experimentation-production parity because of use of standardized frameworks, development of common pre-processing, feature engineering packages and so on. Further, closer collaboration between data scientists and engineers to arrive at shared understanding of non functional serve requirements also helps close the gap between train and serve code.

- AI Models are Rarely Perfect on Day-One. Plan for Continous Improvements: AI models are not static, they need to improve or adjust to data over time. In order to improve the model it is important to have access to data that is representative of real data the model is getting used on. In traditional software projects, limited exception and error logging is done in production. The main reason for logging is to help developers debug any production issues that might arise. However, in AI implementations it is important to have a strategy to collect payloads, as they are the real examples of data the model is being used for. Payload data needs to be brought back into the train pipeline to improve the model. Once more training data is available data scientists are again required to go through the data through train pipeline to arrive at improved model, followed by engineering team who needs to optimize for performance and deploy. This makes model improvement a recurring and continuous process.

- Automate the Train Pipeline: Having automated training pipelines can help significantly reduce the time a data scientist has to spend in improving model. When new training data comes in, the train pipeline would be executed, and as part of this, multiple experiments are auto-executed. Data scientists can then select the best model and push it for deployment. Best practices and tooling for continous integration and delivery from traditional software development life cycle (SDLC) can help reduce engineering time spent in deploying a new model.

Organizations that rely on AI models as part of their daily operations have made significant progress in maturing their train pipelines [18, 19, 20]. New tools to manage train and serve pipeline are regularly being released in market, e.g. version manage AI projects [21], integrated environments to build and run AI models [22].

## VII. TEST PIPELINE

Testing is an investigation process conducted to derive insights about the quality of the machine learning models under test. Here, we share some of the best practices in testing based on our experience.

- Be Prepared to Iterate between Train and Test: While we often have lots of choices to learn and apply various machine learning algorithms on our data sets, selecting the final best model out of many good working models is a challenging and time-consuming task. In practice, train data scientists and testers often work together to compare the performance of models generated with different algorithm parameters before deciding which parameters to use; they may also compare performance of the models using different feature-based representations to ensure the selected features are improving the models as expected.

- You can Test All You Want but Real-world Can Still Shock You! Don't Judge a Machine Learning Model in First Iteration: The most common way to test model performance in machine learning is to look at summary statistics such as Precision, Recall, Accuracy, F-measure, Micro F1-scores and Macro F1-scores, Mean Absolute Error, etc. Another widely used measure for model testing is the confusion matrix which contrasts model predictions against ground truth labels in a table of aggregated values. Typically, testing has been mostly done on fixed and predefined datasets, providing a limited testing coverage. As a result, the quality of machine learning models can vary widely when tested with real-world test cases depending on how close the real-world test cases are to the pre-defined test datasets. Therefore, having a clear idea that the model can improve overtime with data from the real-world is important to keeping things in perspective.

- Testing is Not Just a One-time Build Activity. It is Continous Throughout an AI model's lifecycle. Keep the Test Datasets Updated: In AI services, there is a notion of continuously improving the accuracy of the models as more data becomes available either via continuous data acquisition process or from payload data. While each iteration of the machine learning model can be tested on the same set of standard datasets it can be unfair to test systems on only one set when the newer models have 'seen more of the world' via more training data. As more and more training data is added from different sources, testing should be an iterative and dynamic process wherein test cases are continuously updated to improve the test coverage to represent the new world they live in. This makes comparing models from one version to another difficult. There is no perfect solution for this. We have noted that maintaining old and new test cases and testing model versions on all test cases each time gives a comprehensive view of the quality of the current and past models.

- Whose Side is the Real 'Truth'? Sometimes Machine Learning Models Are Both Right and Wrong! The 'ground-truth' can be different for different people in certain domains. For example, what appears as a complaint to some may appear as a neutral statement to others. Therefore, user acceptance testing of AI-based services may depend on individual user perceptions. Special user perception testing needs to be instituted in addition to conventional performance testing in cases where ground truth can be ambiguous. As the predictions of models from one version to another can often be different, such user perception testing has to be done continuously to allow testers to select the best user perceived model in some cases.

- Scenario Testing for Applications Consisting of Multiple Machine Learning Models: Deploying a machine learning model in real-world often requires fitting specific test scenarios. Exhaustive testing may not possible due to large number of data combinations and large number of possible use cases. Scenario testing is to make sure that end to end functionality of a model under test is working as expected. Tester needs to check and perform the action as how end users are using application under test. Tester often needs to consult the client, stakeholder or developers to prepare suitable scenarios.

- Adversarial and Long Tail Testing: A mature organization needs to do proactive testing for understanding and guiding effective AI model testing. Proactive testing differs from conventional testing metrics in two aspects. First, it extends the coverage of the testing dataset by dynamically collecting supplementary data. Second, AI developers can collect additional data belonging to certain categories to target corner cases. To create failed cases at scale, adversarial sample has attracted attention in machine learning communities in recent years. For example, different perturbation strategies (e.g., insertion, deletion, and replacement) have been proposed to evade DNN-based text classifiers [23].

VIII. MODEL FAIRNESS, TRUST, TRANSPARENCY, ERROR ANALYSIS AND MONITORING

With the increasing adoption of AI models in real-world, comes the need for fairness, trust and transparency. In some industries such as insurance, and financial services, the need for these is further accentuated by the legal and regulatory requirements. An AI model that simply makes a prediction without explaining why it arrived at that prediction may not be acceptable in certain use cases. Each of the topic of fairness, trust and transparency must be dealt with seperately.

Set Proper Goals for AI Models to Mitigate Undesirable Biases and Start with Test Cases: Statistical machine learning models rely on biases in data to learn patterns. Therefore, the concept of data bias by itself is not bad. What people mean, when they say biases is 'undesirable biases'. We argue that undesirable biases creep in because of lack of discipline in setting proper goals for the AI models. Proper goals can be set for AI models by preparing test cases upfront and setting specific objectives on what is expected of the model. As noted in the data requirements section, asking data scientists to 'build a world-class face recognition AI model' is too broad, vague and leads to unanticipated biases. A more specific and focused goal such as: 'build a face recognition service that can detect male/female genders, with pre-defind specific age groups, and these specific subset of races, and ethnicities in the requirements document (which is grounded in a standard taxonomy from a neutral organization such as the United Nations Race and Ethnicity taxonomy)) with at least 90% accuracy on '*these*' given specific test datasets' where 'these' test datasets were carefully crafted by the offering management ream to have an even distribution of all the genders, age groups, specific races, and ethnicities for which the model is supposed do well. That is a specific, focused and measurable goal that a data scientist can build a model for. Such a focused goal is also non-disputable, measurable and tested for biases. It is this lack of specificity that leads to undesirable biases.

Declare Your Biases to Establish Trust: Rarely do organization have unlimited budgets and time to collect representative samples to prepare most comprehensive datasets that can avoid undesirable biases completely. One can, at best, mitigate biases with careful planning. Therefore, we'd argue that it is more practical for a machine learning model to declare its biases than to pretend that it is unbiased or that it can ever be fully unbiased. That is, offering managers must declare what the model is trained on. That way, the consumers of the model know exactly what they are getting. This establishes trust in AI models. This is akin to having nutrition labels on processed and packaged foods. People can judge based on the contents, whether a particular snack item is right for them or not. While not all machine learning model builders may have the incentive to declare the secrets of their ingredients, it may be required in some regulated industries.

Do We Always need full explainability? Let the use case drive the needs and select machine learning algorithms accordingly: We still don't know why and how certain medicines work in human body and yet patients rarely question when a doctor prescribes a medicine. They inherently the trust the doctor to give them the best treatment and trust their choice of medicine. Citing such analogies, some argue whether full explainability may not be always needed. Whether or not the medical analogy is appropriate for a business domain, one thing is clear. Some use cases demand full transparency while others are more forgiving. For example, a sentiment prediction model which aims to predict consumer sentiments against products from social media data may not need the same level of transparency as a loan approval AI model which is subject to auditability. Therefore, based on the use case and need, AI model development team must set transparency goals ahead of time. A data scientist training an AI model can use these requirements in making the right kind of AI model that might offer more explainability or not.

Diagnose Errors at Scale. Traditionally, error analysis is often manually performed on fixed datasets at a small scale. This cannot capture errors made by AI models in practice. A mature error analysis process should enable data scientists to systemically analyze a large number of "unseen" errors and develop an in-depth understanding of the types of errors, distribution of errors, and sources of errors in the model.

Error Validation and Categorization. A mature error analysis process should be able to validate and correct mislabeled data during testing. Compared with traditional methods such as Confusion Matrix, a mature process for an organization should provide deeper insights into when an AI model fails, how it fails and why. Creating a user-defined taxonomy of errors and prioritizing them based not only on the severity of errors but also on the business value of fixing those errors is critical to maximizing time and resources spent in improving AI models.

Continously Monitor Models for Drift: Statistical properties of the target variable, which the model is trying to predict, may change over time in unforeseen ways". This is called *drift*. For example, users' preferences or sentiments may change somewhat rapidly in certain cases thereby making a recommendation model which was trained using historical preferences to be no longer relevant in predicting in current preferences. Therefore, models have to watch for such shifts in target variables. In such cases, model needs to be updated. That often involves collecting new ground truth data, verification of old ground truth data and change as required, and retraining the model. A mature organization which develops such machine learning model needs to have a way to detect such change of model behavior using well defined metrics. This may involve regularly collecting test data from varieties of sources, testing the model output using such data, and identifying when there is significant change of accuracy.

Continuously Monitor Models for Misalignments: Continuous update of the model with the addition of new data also needs some caution. A machine learning model is trained to achieve a specific goal, and continuous update may cause mismatch from that goal. For example, a machine learning model may be originally trained to predict sentiment from short text, e.g., Tweets. However, once it is deployed, some users may try to use the model to predict sentiment from customer-agent conversations on call logs. Since the model was originally not trained on conversational data, it may fail more often for such data. Model developers may find those failed cases in payload logs and may retrain model with such data. However, this changes training data distribution and may make the model perform poorly on Tweets, which was the original goal. Similarly, continuous update of the model may change label distribution which could cause shift from the original goal. Each time, a model is being updated, an organization developing such model needs to assess whether such update causes misalignment from the original goal. They need to have tools to test this.

Curate Training Data for Specific Business Needs and Continuously Learning. A mature organization should be able to detect weakness in training data that has been run against the model. This enables the AI Operations lead to gain insights on which data would be helpful, harmful, or repetitive for certain use cases. This in turn, drives optimization such as lower resource cost to curate high-quality training data that is representative of market needs and continuously improve the model.

Version Models and Manage their Lineage to Better Understand Model Behavior Over Time: An organization may have multiple versions of a machine learning model. A mature organization needs to maintain different versions in a data-store. They should also keep the lineage of training data used to building such models. In addition, they should be able to run automated tests to understand the difference between such models using well defined metrics, and test sets. With each version, they should track whether model quality is improving for such test sets.

In conclusion, we have presented a set of best practices appliable to building and managing the life cycle of machine learning models. Appendix A contains snippets of our framework for select stages of the pipeline. We are unable to publish the full framework in this paper due to space limitations. However, we intend to make it available on our company website for reference.

## IX. CONCLUSIONS AND DISCUSSION

The field of Artificial Intelligence (AI) is seeing resurgence spurred by the latest advances in algorithms, the availability of the latest algorithms in various software libraries, improvements to compute power, and easy access to that compute power via Cloud. AI powered applications are being actively explored, developed and deployed by companies around the world. In this paper we argued that traditional software development lifecycle methodologies fall short when managing AI models as AI lifecycle management has many differences from traditional software development lifecycle. This necessitates a framework for managing the life cycle of these AI models in business applications. We articulated the lifecycle of an AI model, presented best practices in each lifecycle stage of an AI model from our experience of building practical, large scale, AI models. We also presented a re-interpretation of software capability maturity model (CMM) for AI mode lifecycle management.

Starting with various new roles that are involved in building and managing AI models such as data managers, train lead, AI operations management lead, and various new life cycle steps such as data pipeline management, feature preparation, model training and operations management, AI model development, and management are posing interesting new and unique challenges to organizations exploring and deploying production ready AI models. On the one hand, organizations are excited about the possibility of untapping the potential of unstructured data sitting by the wayside in their enterprises, which amounts to over 80% of the data that is generated by enterprises today by a recent study, with AI models. On the other hand, organizations are having to adjust their expectations as the initial set of AI models they deployed are performing below expectations. The prediction accuracy of the initial AI models deployed were a disappointment for organizations. There are many reasons for this mismatch of expectations.

First, organizations did not realize that AI models are probabilistic models and that they learn patterns from messy data and it is not always possible to model the patterns in data exactly to fit every data point. That the accuracy will not be 100% takes a big expectation adjustment for organizations still. To address this, more education is needed in organizations. Overtime, AI models will get better by learning from more and more data but initial iterations must be treated as such.

The second reason why the initial iterations of AI models often do not meet organizations' expectations is that these models are usually built on general purpose data. We call these models that are built on general purpose data as *base* models. While a *base* model provides foundations for an AI model, industry/domain adaptation and customer adaptation are needed for achieving viable and usable AI models in specific domains, specific use cases on specific data types. Each company has its own vocabulary, policies, products, offers and terminology. When the AI models are given the benefit of learning from this industry specific and use case specific data on top of the *base* models, the models are likely to do a much better job of predicting things more accurately. We presented two ways to address this problem: a) by designing AI models with customization as the one of the primary design points for continuous improvements. AI models that can't be customized or those that don't expose retraining either via APIs or via services engagements remain static and not very useful as they cannot learn continuously to improve and b) consciously planning the AI model development pipeline for continuous improvement and iterations. We presented several best practices for achieving this.

Another reason for hesitation in adopting AI models is that organizations find them to be black-boxes and non-transparent. This is especially true for models trained with deep learning techniques. In regulated industries, ability explain the behavior of AI models is extremely important. Therefore, AI models and tools must focus on addressing fairness, trust and transparency topics. In this paper, we have presented various best practices and a maturity assessment framework for addressing these topics.

Implementing these best practices requires many innovations, tools and techniques. So much AI is needed throughout the AI lifecycle management to build and management AI models. We are excited about the research and innovation possibilities and frontiers that this offers. A journey informed by best practices and maturity awareness is the best way to get there.

## ACKNOWLEDGMENTS

We would like to acknowledge our IBM colleagues Pierre Arnoux, Neil Boyette, Hau-wen Chang, Haibin Liu, Sudhir Koka, Michael Picheny, Ruchir Puri, Beth Smith, Samuel Thomas, Olivia Buzek, Pawan Lakshmanan, Laura Chiticariu, Padma Malladi, Majid Irani, Yuankun Song, Yingbei Tong, Gary Diamanti, Inkit Padhi and Kaveh Massoudian as additional contributors to this work whose names could not be accommodated in the main author list.

**Appendix A**. A snippet of our Machine Learning Maturity Framework is attached below. A more detailed one could not be attached due to space limitations but will be made available upon request or posted on the company website shortly.

| Capability | Detailed Capability | Role | Initial | Repeatable | Defined | Managed | Optimizing |
|---|---|---|---|---|---|---|---|
| AI Model Goal Setting | Target Scope and Purpose | Offering Manager | * Scope is unclear - goal is set to statements as vague as 'build world-class models'. For what, for which industries, for whom, for which type of data is not enunciated. | * Qualitative goal setting. Build models with 90% accuracy on Sentiment analysis for any type of data. No idea if 90% is achievable or not. * Qualitative scope and targets. Still lacks clarity, and formality. | * Acknowledgement of the importance of scope clarity across the organization. * Some clarity starts to emerge on use cases, industries, and users needs and perspectives are taken. | * Proactively scopes the model goals and targets after each iteration and adjusts and communicates model purpose to clients and users. * All parties have a clear understanding of what to expect of the model. | * Model's purpose, goals are clearly stated. Methods, automation tools are in place to measure and monitor the model metrics. * Explictly declares what training data the model is built with and declares known biases with clients to establish trust. * As much clarity as what the model isn't meant to do as there is clarity on what the model is meant to do. |
| Data Pipeline Management | Data Annotation | Data Manager | * Have initial annotation guidelines | * Have consistent annotation procedures across annotation teams. | * Have a process to define the difficulty of the annotation tasks and split the tasks between subject-matter experts and non-expert workers for data annotation accordingly. | * have a routine feedback loop between data scientists and data annotators. * Have machine learning methods in the pre-annotation process to reduce the human annotation efforts. * Manage the workload between the human annotation and machine annotation. | * Have a process to optimize the cost and annotation outcome. * update annotation guidelines and produces timely and appropriately based on new requirements from AI models. |
| Feature Preparation Pipeline | Externalized feature processing pipeline | Train Lead | * Feature processing pipeline is not designed as an indepnt module. Part of feature processing appears in the data pre-processing/preparation. * Train and test pipelines might not be following exact same steps or order to generate features | * Feature processing is an indepnt module, but not exetrnalized seperately. Hence, same feature processing module code is repeated train and test pipelines. | * Feature processing module is externalized and shared between train and test. Hence code maintainance is extremely efficient. * The cost of the externalization on the reponse time is not considered | * The cost of the externalization on the reponse time is throughly understood. * However, extrenalization cost may have come at the cost of removing the important features. | * Altenative strategies have been employed, inlcuding efficient and faster implementation of feature extraction module, to make sure that cost of externalization is minimal. |
| Train Pipeline Management | Modularizing Train code | Train Lead | * Left to data scientist | * Have basic guidelines for data scientist to create separate modules for feature engineering, training. | * Data science team publishes, discusses best practices. | * Train code and model pipeline review process established. | * Shared repository of reusable feature extraction modules, and training frameworks. * Data scientists routinely experiment with different feature sets, training modules to build better models. |
| Test Pipeline management | Test Metrics and Goals | Test Lead | * No metrics or goals were set for testing. | * Only test on summary metrics such as Precision, Recall, Accuracy, F-measure. * Goal is set to merely improve the accuracy of the model. | * Multiple test goals are manunally created by test lead. * Multiple test metrics are mannually selected by test lead. | * Test goals are created to address specific concerns of the model * Test metrics are defined and refined for a variety of coverage. | * Test goals are created and customized for domain-specific problem with high market viablity * Test metrics reflect the domain and scope of where the model will be deployed |
| Model Quality, Performance and Model Management | Quality metrics measurement | Quality Assurance Lead | * non-existent | * Manually looking at model output for few test cases and assessing for goodness | *Well defined metric set for measuring model quality | *Automated means established to measure model quality using well defined metirc | *Tools developed to measure model quality using well defined metric |
| Model Error Analysis | Error Categorization | Quality Assurance Lead | * Have single-score metrics such as precision and recall | * Have traditional error categorization methods such as Confusion Matrix | * Have a process to define the testing covering of exing datasets. * Have a process to define user-defined error taxonomy. | * Have an automated process to categorize errors into severity levels can help an organization to prioritize their efforts to address critical errors in their AI applications | * Have an optimized process to prioritize errors in terms of user-defined error types, data-driven metrics, and error severity. *Have a process to automate the classification of errors into the user-defined error types. * Categorize Errors based on data-specific features. For example, for a text-based model (e.g. Nature Language Understanding), errors made by the model can be categorized based on text topics, sentence length, and etc. |
| Model Fairness & Trust | Undesirable bias awareness | AI Operations Manager | * Not aware of the existence of any type of bias. | * Only aware the existence of a set of pre-defined explicit bias (e.g. gender, age, race). * Do not aware of the existence of explicit bias out side of the pre-deifned list, as well as implicit bias. | * Aware the existence of a broad set of explicit bias, not just limited to the pre-defined list. * Do not aware of the existence of implicit bias. | * Aware the existence of both explicit and implicit biases. | * Aware the existence of both explicit and implicit biases. * Being able to automatically quantify the risks of bias exposure, and to inform the stakeholders accordingly. |
| Model Transparency | Awareness of Explainability of algorithmic choices | AI Operations Manager | * Not aware the importance and necessity of model explainability | * Aware the importance and necessity of model explainability. | * Aware the importance and necessity of model explainability. * Provide only some global level interpretation of the model (e.g. training information, underlying algorithms, performance description). | * Provide both global level (e.g. training information, underlying algorithms, performance description) and local level (contributing features regarding one specific case)interpretation of the model. * Being able to adjust the mode interpretations based on the different background of end users (e.g. technical vs. non-technical, novice vs. expert). | * Allow users to interact with the model interpretation and to flag attributes that they think the model learned incorrectly. * Being able to adjust the model according to the user's feedback. * Allow users to appeal to certain model prediction. |